\documentclass{article}

\usepackage[preprint]{neurips_2025}

\usepackage[utf8]{inputenc} 
\usepackage[T1]{fontenc}    
\usepackage{hyperref}       
\usepackage{url}            
\usepackage{booktabs}       
\usepackage{amsfonts}       
\usepackage{nicefrac}       
\usepackage{microtype}      
\usepackage{amsmath}
\usepackage{makecell}
\usepackage{graphicx}
\usepackage{adjustbox}
\usepackage{tabularx}
\usepackage{multirow}

\usepackage[table]{xcolor}

\title{NbBench: Benchmarking Language Models for Comprehensive Nanobody Tasks}

\author{%
  Yiming Zhang \\
  Department of Computational Biology and Medical Sciences \\
  The University of Tokyo \\
  Japan\\
  \texttt{zhang\_yiming\_25@stu-cbms.k.u-tokyo.ac.jp} \\
  \And
  Koji Tsuda\thanks{corresponding author} \\
  Department of Computational Biology and Medical Sciences \\
  The University of Tokyo \\
  Japan \\
  \texttt{tsuda@k.u-tokyo.ac.jp} \\
}

\begin{document}

\maketitle

\begin{abstract}
Nanobodies—single-domain antibody fragments derived from camelid heavy-chain-only antibodies—exhibit unique advantages such as compact size, high stability, and strong binding affinity, making them valuable tools in therapeutics and diagnostics. While recent advances in pretrained protein and antibody language models (PPLMs and PALMs) have greatly enhanced biomolecular understanding, nanobody-specific modeling remains underexplored and lacks a unified benchmark. To address this gap, we introduce NbBench, the first comprehensive benchmark suite for nanobody representation learning. Spanning eight biologically meaningful tasks across nine curated datasets, NbBench encompasses structure annotation, binding prediction, and developability assessment. We systematically evaluate eleven representative models—including general-purpose protein LMs, antibody-specific LMs, and nanobody-specific LMs—in a frozen setting. Our analysis reveals that antibody language models excel in antigen-related tasks, while performance on regression tasks such as thermostability and affinity remains challenging across all models. Notably, no single model consistently outperforms others across all tasks. By standardizing datasets, task definitions, and evaluation protocols, NbBench offers a reproducible foundation for assessing and advancing nanobody modeling. All datasets and source code used in our benchmark are publicly available at \url{https://github.com/ZHymLumine/NbBench}.

\end{abstract}

\section{Introduction}
Antibodies are essential components of the adaptive immune system, capable of recognizing and neutralizing diverse antigens such as viruses, bacteria, and toxins~\cite{lu2020development}. This high specificity has made antibodies indispensable in therapeutics, diagnostics, and molecular biology~\cite{nelson2010development}. Structurally, conventional antibodies are composed of two heavy and two light chains forming a Y-shaped molecule, with the antigen-binding regions located within the variable domains~\cite{schroeder2010structure}.

\textit{Nanobodies} (also known as \textit{VHHs} or \textit{single-domain antibodies}) are the antigen-binding domains of heavy-chain-only antibodies found in camelids~\cite{el2025nanobody}. Unlike conventional antibodies, nanobodies consist of a single variable domain and lack light chains, yet retain high binding affinity and specificity~\cite{muyldermans2013nanobodies, jovvcevska2020therapeutic}. Their simplified architecture confers several advantages: small size (~15 kDa), high solubility, strong thermal and chemical stability, and the ability to access hidden epitopes~\cite{el2025nanobody, DeMeyer2014NanobodybasedPA, van2016nanobodies}. These features make nanobodies attractive for applications in molecular imaging, therapy, and synthetic biology~\cite{alexander2024discovery, bao2021nanobody}.

Despite these advantages, nanobodies pose distinct modeling challenges. Their lack of paired light chains, longer and more flexible CDR3 loops~\cite{muyldermans2013nanobodies}, and smaller structural databases make it difficult to apply conventional antibody modeling techniques~\cite{steeland2016nanobodies}. While large datasets like the Observed Antibody Space (OAS)\cite{kovaltsuk2018observed, olsen2022observed} exist for conventional antibodies, nanobody-specific resources such as INDI\cite{deszynski2022indi} and VHHCorpus-2M~\cite{tsuruta2024sars} remain relatively limited.

Recent advances in protein language models (PLMs), such as ESM~\cite{lin2023evolutionary}, ProtBert~\cite{elnaggar2021prottrans}, and MSA Transformer~\cite{rao2021msa}, have demonstrated strong performance on structure and function prediction tasks~\cite{rao2019evaluating, madani2020progen, meier2021language}. These models leverage large-scale self-supervised training on protein sequences to learn contextual representations that capture structural and evolutionary information. Building on this, antibody-specific language models (PALMs) like AntiBERTy~\cite{ruffolo2021deciphering}, AbLang~\cite{olsen2022ablang}, AntiBERTa2~\cite{barton2024enhancing}, and IgBert~\cite{kenlay2024large} have been developed to model immunoglobulin sequences with higher accuracy. However, their applicability to nanobody modeling remains largely unexplored, as they are predominantly trained and evaluated on conventional antibodies.

More recently, nanobody-specific models such as NanoBERT~\cite{hadsund2024nanobert}, NanoBERTa-ASP~\cite{li2024nanoberta}, DeepNano~\cite{deng2024nanobody}, and VHHBERT~\cite{tsuruta2024sars} have emerged. However, these models are typically developed for narrow tasks and use inconsistent datasets and evaluation metrics, making fair comparisons difficult. Furthermore, most prior work does not assess generalization to unseen nanobody types or functions.

To address these limitations, we introduce \textbf{NbBench}, the first comprehensive benchmark suite for nanobody modeling. NbBench spans eight biologically relevant tasks across nine curated datasets, covering structural annotation, sequence generation, function prediction, and developability assessment. It provides standardized task definitions, data splits, and evaluation metrics to facilitate fair, reproducible comparisons between general-purpose, antibody-specific, and nanobody-specific language models. Using NbBench, we evaluate eleven pretrained models and analyze their generalization performance across task types, revealing key insights into the capabilities and limitations of current language models for nanobody modeling.

Overall, our contributions can be summarized as follows:
\begin{itemize}
\item We introduce \textbf{NbBench}, the first comprehensive benchmark for nanobody modeling, spanning eight biologically meaningful tasks across nine curated datasets.
\item We systematically evaluate eleven pretrained protein, antibody, and nanobody language models, revealing their generalization capabilities and failure modes across structure, function, and developability tasks.
\item We establish a standardized evaluation protocol for nanobody tasks, enabling fair model comparison and providing actionable insights for future model development and deployment.
\end{itemize}

\begin{figure}[htbp]
    \centering
    \includegraphics[width=\textwidth]{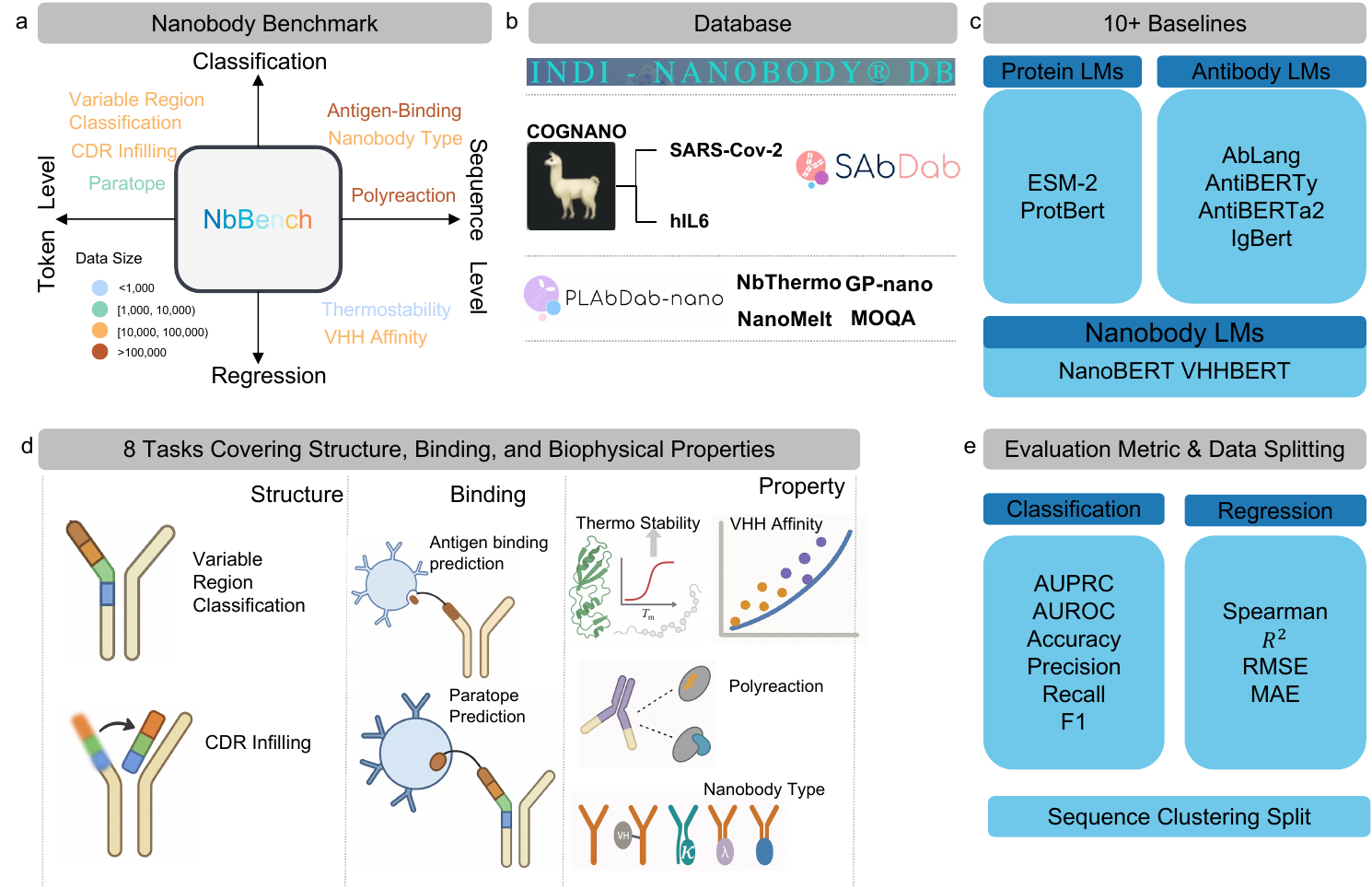}
    \caption{
    \textbf{Overview of NbBench.}
    (a) Eight tasks spanning token- and sequence-level classification and regression.  
    (b) Data sources include INDI, SAbDab, PLAbDab-nano, and task-specific datasets.  
    (c) 10+ baselines from pretrained protein, antibody, and nanobody language models.  
    (d) Visual representations of tasks grouped into structure understanding, binding, and biophysical and functional property prediction.  
    (e) Standard evaluation metrics and clustering-based data splits.
    }
    \label{fig:benchmark-overview}
\end{figure}

\section{Related Works}
\textbf{Pretrained Protein and Antibody Language Models (PPLMs and PALMs).}Protein language models such as ESM-2~\cite{lin2023evolutionary}, ProtBert~\cite{elnaggar2021prottrans}, and MSA Transformer~\cite{rao2021msa} have demonstrated strong transfer performance on structure and function prediction tasks~\cite{rao2019evaluating, madani2020progen, meier2021language, chen2023structure, zhang2025hyena}. These models leverage large-scale self-supervised learning and evolutionary context to build general-purpose protein representations. In parallel, antibody-specific language models (PALMs) have been introduced to model immunoglobulin sequences more effectively. AntiBERTy~\cite{ruffolo2021deciphering} was the first such model trained on 588 million heavy/light chain sequences from the OAS database, revealing that antibodies can be embedded along trajectories resembling affinity maturation. AntiBERTa~\cite{leem2022deciphering} was fine-tuned for paratope prediction, while AbLang~\cite{olsen2022ablang} trained separate models on heavy (AbLang-H) and light (AbLang-L) chains, outperforming ESM-1b in recovering masked residues. BERT-DS~\cite{porebski2024rapid} and AntiBERTa2~\cite{barton2024enhancing} further expanded training datasets to tens or hundreds of millions of antibody sequences, with AntiBERTa2 incorporating multimodal contrastive learning between sequence and structure. IgBert~\cite{kenlay2024large}, initialized from ProtBert, was trained on over 2 million paired antibody sequences. EATLM~\cite{wang2023pre} introduced evolution-guided pretraining objectives, such as germline ancestor prediction and mutation site masking, and also proposed the ATUE benchmark. Despite this rapid progress, most PALMs have focused on conventional antibodies and are primarily evaluated on limited downstream tasks such as paratope prediction, sequence recovery, or general binding classification.

\textbf{Nanobody modeling and current limitations.}While antibody-specific language models have advanced rapidly, their applicability to nanobody modeling remains largely unexplored. Nanobodies—single-domain fragments derived from camelid antibodies—have emerged as promising tools for diagnostics, therapeutics, and molecular imaging~\cite{salvador2019nanobody, alexander2024discovery, bao2021nanobody}. Their compact size and ability to bind hidden epitopes make them attractive alternatives to conventional antibodies~\cite{salvador2019nanobody, liu2025unveiling}. Classical in silico approaches—such as docking, molecular dynamics, and epitope mapping—have long been used to analyze nanobody-antigen interactions~\cite{el2025nanobody, joubbi2024antibody}, but are often resource-intensive and lack scalability. Structure prediction tools like AlphaFold2~\cite{jumper2021highly} and OmegaFold~\cite{wu2022high} improve modeling accuracy, yet still rely on structural inputs or evolutionary features~\cite{valdes2023structural}. Although a few deep learning models such as NanoBERT~\cite{hadsund2024nanobert}, NanoBERTa-ASP~\cite{li2024nanoberta}, and VHHBERT~\cite{tsuruta2024sars} have been proposed for thermostability, paratope, or binding prediction, each is designed for a narrow task and evaluated in isolation. A unified framework for evaluating nanobody modeling remains missing.

\textbf{Benchmarking efforts and the need for NbBench.}Large-scale benchmarks have played a critical role in evaluating protein and antibody language models. TAPE~\cite{rao2019evaluating} introduced five foundational tasks, such as secondary structure and stability prediction, to assess general-purpose protein representations. PEER~\cite{xu2022peer} expanded to multi-task evaluation across sequence- and token-level objectives, while FLIP~\cite{dallago2021flip} focused on mutational fitness landscapes to benchmark protein engineering. More recently, ATUE~\cite{wang2023pre} proposed the first antibody-specific benchmark, covering four real-world tasks—antigen binding prediction~\cite{mason2021optimization}, paratope prediction~\cite{liberis2018parapred}, B cell maturation analysis~\cite{mroczek2014differences}, and SARS-CoV-2 antibody discovery~\cite{raybould2021cov,zaslavsky2022disease}—designed to assess the biological understanding and evolutionary awareness of pre-trained antibody models. Despite these advances, all existing benchmarks share several limitations in the context of nanobody modeling. First, they primarily focus on general proteins or conventional antibodies, with little attention to the structural and functional uniqueness of nanobodies. Second, tasks are often narrow in scope and evaluated in isolation, preventing consistent comparison across models. Third, none of the benchmarks provide a unified evaluation across the diverse range of nanobody-relevant tasks such as structure modeling, binding function, and developability. To address these gaps, we introduce \textbf{NbBench}, the first multi-task benchmark suite specifically designed for nanobody modeling. NbBench standardizes datasets, metrics, and splits across eight biologically meaningful tasks, enabling systematic and fair evaluation of both general-purpose and antibody-specific language models in nanobody-centric scenarios.

\section{Benchmark Tasks}
\label{headings}
NbBench consists of 8 tasks across 9 datasets grouped into three categories: structure and region understanding, binding prediction, and biophysical property prediction. These tasks cover token-level and sequence-level classification, regression, and sequence generation with an overview provided in Table~\ref{tab:nbbench-datasets}.
\subsection{Structure and Region Understanding}

\textbf{Variable Region Classification (VRCls)} classifies each residue in a nanobody sequence into structural regions: framework regions (FRs) or one of the three complementarity-determining regions (CDR1, CDR2, CDR3). This task is formulated as an amino acid-level classification problem, where the model must predict a categorical label for each amino acid in the sequence based on its structural and functional role. We collect data from the INDI database~\cite{deszynski2022indi}, using the \texttt{abgenbank}, \texttt{manual}, and \texttt{structure} subsets for training, and the \texttt{patent} subset for testing. We remove test sequences that are exact duplicates of those in the training set, and ensure that the average sequence similarity between training and test sets remains at 75\%.

\hspace*{1em}\textit{Evaluation metric}: accuracy, precision, recall, and F1 score.  

\hspace*{1em}\textit{Biological impact}: The three complementarity-determining regions (CDRs) are the primary determinants of antigen binding, while the surrounding framework regions (FRs) maintain the structural integrity required for proper presentation of the CDRs. However, the activity of a nanobody is not only determined by the CDR sequences themselves; the three-dimensional structure and interactions between all variable regions, including FRs and CDRs, play a crucial role in antigen recognition and binding~\cite{nakanishi2008critical}. Therefore, accurately classifying every residue in the variable region—whether it belongs to a framework or a CDR—is essential. This comprehensive annotation enables more effective downstream applications such as CDR grafting, humanization, affinity maturation, and structure-aware generative design, and improves the interpretability and reliability of nanobody models.

\textbf{CDR Infilling (CDRInf)} reconstructs masked CDR regions within a nanobody sequence. Given a sequence where the CDRs are masked or removed, the model is expected to generate plausible CDR sequences that complete the nanobody. This task is formulated as a token-level sequence generation problem.
The dataset and train-test split are consistent with the CDR classification task. We use the same subsets from INDI and apply the sequence similarity-based splitting strategy.

\hspace*{1em}\textit{Evaluation metric}: Exact match accuracy (hard recovery) and similarity score based on BLOSUM62 (soft recovery).  

\hspace*{1em}\textit{Biological impact}: CDR infilling evaluates the model’s ability to generate diverse yet functional antigen-binding regions within a fixed nanobody scaffold. Success on this task suggests a capacity to learn structure-function relationships that could be used in de novo nanobody design, especially in scaffold-constrained generative pipelines or CDR-optimization loops.

\subsection{Binding Prediction}
\textbf{Antigen-Nanobody Binding Prediction (AgNbBind)} predicts whether a nanobody can bind to a given antigen. This is formulated as a binary classification task. We include two separate datasets, each corresponding to a different antigen: (1) SARS-CoV-2~\cite{tsuruta2024sars} and (2) human IL-6 (hIL6)~\cite{tsuruta2023avida}.

\hspace*{1em}\textit{Evaluation metric}: AUROC (area under the receiver operating characteristic), AUPRC (area under the precision-recall curve), accuracy, precision, recall and F1 score. 

\hspace*{1em}\textit{Biological impact}: Predicting binding interactions is critical for therapeutic antibody discovery. It helps prioritize candidates with high affinity and broad neutralization potential, and accelerates lead selection and optimization.

\textbf{Paratope Prediction (Paratope)} identifies which residues in the nanobody are directly involved in antigen binding. This is framed as a sequence labeling task, where the model predicts a binary label for each residue in the nanobody, indicating whether it is part of the paratope.

\hspace*{1em}\textit{Evaluation metric}: AUROC, AUPRC, accuracy, precision, recall and F1 score.  

\hspace*{1em}\textit{Biological impact}: Understanding which residues form the paratope reveals how nanobodies interact with antigens. This insight supports rational nanobody engineering, such as improving binding affinity through site-directed mutagenesis or redesigning specific binding loops.

\subsection{Biophysical and Functional Property Prediction}

\textbf{Thermostability Prediction (Thermo)} estimates the melting temperature (T$_m$) of nanobody sequences, which reflects their thermal stability. This task is framed as a regression problem, where the model is trained to predict a continuous T$_m$ value from the nanobody sequence alone. We collect thermostability data from multiple curated sources, including \textit{NbThermo}~\cite{valdes2023nbthermo}, \textit{TEMPRO}~\cite{alvarez2024tempro}, and \textit{NanoMelt}~\cite{ramon2025prediction}, which report experimentally measured or computationally inferred T$_m$ values for diverse nanobodies. We apply sequence similarity-based splitting to ensure low redundancy between training and test sets, and to simulate generalization to novel scaffolds.

\hspace*{1em}\textit{Evaluation metric}: Spearman correlation, $R^2$, RMSE (root mean square error), and MAE (mean absolute error). 

\hspace*{1em}\textit{Biological impact}: Thermostability is a key developability attribute for therapeutic and diagnostic nanobodies. Higher T$_m$ values are associated with better solubility, longer shelf life, and resilience in extreme environments, making thermostability prediction an important task for early-stage candidate screening and protein engineering.

\textbf{Polyreactivity Prediction (PolyRx)} predicts whether a nanobody exhibits polyreactivity, defined as nonspecific binding to off-target proteins or biomolecules. This task is formulated as a binary classification problem to distinguish between polyreactive and non-polyreactive sequences. We collect data from two prior studies~\cite{harvey2022silico, zhou2024gp}, which experimentally measured and annotated nanobody polyreactivity using a naïve synthetic camelid antibody fragment library.

\hspace*{1em}\textit{Evaluation metric}: AUROC, AUPRC, accuracy, precision, recall and F1 score.  

\hspace*{1em}\textit{Biological impact}: Polyreactivity is a major liability in therapeutic antibody development, as it can cause off-target effects, reduce specificity, and lead to misleading experimental results. Predicting and minimizing polyreactivity during early-stage screening improves the reliability and clinical developability of nanobody-based drugs. 

\textbf{Nanobody Type Classification (NbType)} categorizes single-chain variable fragments (scFvs) into one of five classes: VHH (variable heavy domain of the heavy chain; camelid-derived), VNAR (variable new antigen receptor; shark-derived), VH (variable heavy chains; human-derived), V$\lambda$ (lambda light chains; human-derived), or V$\kappa$ (kappa light chains; human-derived). This is formulated as a multiclass classification task based on sequence features of the variable domain. We collect VHH and VNAR sequences from PLAbDab-nano~\cite{gordon2025plabdab}, while VH, Vkappa and Vlambda sequences are obtained from ~\cite{ramon2024assessing}

\hspace*{1em}\textit{Evaluation metric}: accuracy, precision, recall, and F1 score.  

\hspace*{1em}\textit{Biological impact}:Correctly identifying antibody types is essential for immune repertoire profiling, especially in mixed or synthetic libraries. Differentiating nanobodies (e.g., VHH and VNAR) from conventional antibodies enables more targeted therapeutic and diagnostic engineering. 

\textbf{VHH Affinity Prediction (Affinity)} estimates the antigen-binding strength of a VHH nanobody based on its amino acid sequence. This is formulated as a regression task, where the output is a continuous affinity-related score derived from high-throughput biopanning experiments.
We collect data from the study by~\cite{tuvcs2024extensive}, which applied deep sequencing to phage-displayed nanobody libraries across multiple rounds of selection. Binding scores are computed based on the relative enrichment of sequences between non-specifically bound phages (N.S. phages) and eluted phages in the 4th round of selection. Specifically, the score for each sequence is defined as:
\begin{equation}
\textit{Score} = \frac{\textit{frequency in eluted phages at 4th round}}{\textit{frequency in N.S. phages at 4th round}}
\end{equation}
\hspace*{1em}\textit{Evaluation metric}: Spearman correlation, $R^2$, RMSE and MAE.  

\hspace*{1em}\textit{Biological impact}: Accurate prediction of nanobody affinity enables virtual screening of large candidate pools, reducing the need for costly wet-lab experiments. This task supports sequence design strategies such as multi-objective optimization.

\subsection{Data Splitting Strategy}
\paragraph{Dataset Splitting via Sequence Clustering.}
To ensure a fair and biologically meaningful evaluation, we adopt a clustering-based strategy for splitting datasets using MMseqs2~\cite{steinegger2017mmseqs2} with a 70\% sequence identity threshold, except for the \textit{Antigen-Nanobody Binding Prediction} task.

We first cluster the entire sequence database such that highly similar sequences (with identity $\geq 70\%$) are grouped together. To encourage diversity, 20\% of the sequences are sampled from low-frequency clusters to form the test set. The remaining sequences are split into 70\% for training and 10\% for validation. For regression tasks such as \textbf{Thermo} and \textbf{Affinity}, we further applied stratified sampling based on binned target values to ensure that the label distribution is consistent across training, validation, and test sets.

To verify the effectiveness of the split, we compute the average pairwise identity between the training and test sets. This ensures that the test set contains sequences that are dissimilar to those seen during training, simulating realistic generalization scenarios and minimizing potential data leakage.

This approach reduces redundancy across splits and prevents data leakage. More importantly, it simulates realistic generalization scenarios in which models encounter novel, non-homologous sequences. This is particularly critical for tasks such as nanobody-antigen interaction prediction, where high sequence similarity may lead to inflated evaluation metrics if not properly controlled.

\paragraph{Antigen-Nanobody Binding Split.}
For the Antigen-Nanobody Binding Prediction task, we use a task-specific split based on known biological variations in the antigens:
\textbf{SARS-CoV-2:} The training set includes wild-type (WT), mild variants (e.g., D614G), conserved regions (S2 domain), and immune escape mutations (PMS). The validation set contains unseen variants from the Kappa and Lambda lineages for hyperparameter tuning. The test set consists of major variants of concern (Alpha, Beta, Delta, Omicron), simulating real-world generalization to novel strains.
\textbf{hIL6:} The training set uses the wild-type antigen; the validation set consists of 5 randomly selected mutants, and the test set includes the remaining 25 mutants.

\begin{table}[htbp]
  \caption{
    Overview of benchmark tasks for nanobody modeling.
    \textbf{Token}-level tasks require a label for each amino acid; \textbf{Sequence}-level tasks assign a single label per nanobody sequence.
    \textbf{Cls} and \textbf{Reg} denote classification and regression tasks, respectively.
    Task suffixes: \textbf{-seq} indicates sequence clustering-based splits; \textbf{-tm} and \textbf{-score} denote stratified splits based on melting temperature and binding affinity scores, respectively.
  }
  \label{tab:nbbench-datasets}
  \centering
  \begin{adjustbox}{width=\textwidth}
  \begin{tabular}{llllll}
    \toprule
    \textbf{Task} & \textbf{\#Train/Val/Test} & \textbf{Type} & \textbf{Granularity} & \textbf{Metrics} & \textbf{Source} \\
    \midrule
    \multicolumn{6}{c}{\textbf{Structure and Region Understanding}} \\
    \midrule
    VRCls & 13,703/3,426/2,847 & Multi-class Cls & Token & Accuracy, Precision, Recall, F1 & INDI~\cite{deszynski2022indi} \\
    CDRInf & 13,703/3,426/2,847 & Multi-class Cls & Token & Exact Match, BLOSUM62 Similarity & INDI~\cite{deszynski2022indi} \\
    \midrule
    \multicolumn{6}{c}{\textbf{Binding Prediction}} \\
    \midrule
    SARS-CoV-2 & 27,606/6,706/38,958 & Binary Cls & Sequence & \makecell[l]{AUROC, AUPRC, Accuracy,\\Precision, Recall, F1} & AVIDa-SARS-CoV-2~\cite{tsuruta2024sars} \\
    hIL6 & 13,020/11,637/449,234 & Binary Cls & Sequence & \makecell[l]{AUROC, AUPRC, Accuracy,\\Precision, Recall, F1} & AVIDa-hIL6~\cite{tsuruta2023avida} \\
    Paratope & 851/139/240 & Binary Cls & Token & \makecell[l]{AUROC, AUPRC, Accuracy,\\Precision, Recall, F1} & SAbDab~\cite{wilton2018sdab} \\
    \midrule
    \multicolumn{6}{c}{\textbf{Biophysical and Functional Property Prediction}} \\
    \midrule
    Thermo-seq & 522/92/147 & Regression & Sequence & Spearman, $R^2$, RMSE, MAE & NbThermo/NanoMelt/TEMPRO~\cite{valdes2023nbthermo, ramon2025prediction, alvarez2024tempro} \\
    Thermo-tm & 396/57/114 & Regression & Sequence & Spearman, $R^2$, RMSE, MAE & NbThermo/NanoMelt/TEMPRO~\cite{valdes2023nbthermo, ramon2025prediction, alvarez2024tempro} \\
    PolyRx & 101,854/14,613/25,007 & Binary Cls & Sequence & \makecell[l]{AUROC, AUPRC, Accuracy,\\Precision, Recall, F1} & Harvey/GP-nano~\cite{harvey2022silico, zhou2024gp} \\
    NbType & 12,647/1,960/3,557 & Multi-class Cls & Sequence & Accuracy, Precision, Recall, F1 & PLAbDab-nano/AbNativ~\cite{gordon2025plabdab, ramon2024assessing} \\
    Affinity-seq & 8,888/1,302/2,547 & Regression & Sequence & Spearman, $R^2$, RMSE, MAE & MOQA~\cite{tuvcs2024extensive} \\
    Affinity-score & 8,915/1,274/2,548 & Regression & Sequence & Spearman, $R^2$, RMSE, MAE & MOQA~\cite{tuvcs2024extensive} \\
    \bottomrule
  \end{tabular}
  \normalsize
  \end{adjustbox}
\end{table}

\section{Benchmark Models}

To comprehensively evaluate nanobody modeling performance, we select a suite of baseline language models covering general protein, antibody, and nanobody-specific pre-training domains. These models vary in architecture, pre-training corpora, and design philosophies. As a practical constraint, we only include models with publicly available weights to ensure reproducibility. Unless otherwise noted, all models are evaluated in a frozen setting using amino acid-level tokenization.We describe each model in detail below and present a summary in Table~\ref{tab:nbbench-models}.

\textbf{Protein Language Models.}
(1) \textbf{ProtBert}~\cite{elnaggar2021prottrans} is a BERT-based model pre-trained on 216 million sequences from UniRef100. It serves as a general-purpose protein language model.
(2--3) \textbf{ESM-2}~\cite{lin2023evolutionary} is a Transformer-based model trained on 65 million sequences from UniRef50. We evaluate two variants: ESM-2 150M and ESM-2 650M. These models help assess the transferability of general protein representations to nanobody tasks.

\textbf{Antibody Language Models.}
(4) \textbf{AbLang-H}~\cite{olsen2022ablang} is a RoBERTa-based model trained on 14 million heavy-chain sequences from the OAS database.
(5) \textbf{AntiBERTy}~\cite{ruffolo2021deciphering} is a BERT-based model pre-trained on 588 million antibody sequences from OAS.  
(6) \textbf{AntiBERTa2}~\cite{barton2024enhancing} is a RoFormer-based model trained on 824 million sequences from OAS and a proprietary dataset.  
(7) \textbf{AntiBERTa2-CSSP}~\cite{barton2024enhancing} is a multimodal variant of AntiBERTa2 that incorporates antibody structure via contrastive sequence–structure pre-training.  
(8) \textbf{IgBert}~\cite{kenlay2024large} is initialized from ProtBert and further trained on 2 billion unpaired heavy\&light chain sequences and 2 million paired antibody sequences.  

\textbf{Nanobody Language Models.}
(9) \textbf{NanoBERT}~\cite{hadsund2024nanobert} is a RoBERTa-based model trained specifically on nanobody sequences to explore mutational space and thermostability.  
(10) \textbf{VHHBERT}~\cite{tsuruta2024sars} is a RoBERTa-based model pre-trained on 2 million camelid nanobody sequences from VHHCorpus-2M. It uses extended positional embeddings to handle long CDR3 loops common in nanobodies.

To ensure fair comparison and isolate the representational quality of pretrained models, all models are evaluated in a frozen setting—i.e., their parameters remain unchanged—and only the final task-specific classifier (e.g., MLP) is trained during fine-tuning. This protocol allows us to assess how well each pretrained representation transfers to downstream nanobody tasks.
\begin{table}[htbp]
  \centering
  \caption{Characteristics of pre-trained protein, antibody, and nanobody language models. ``M'' stands for million.}
  \label{tab:nbbench-models}
  \begin{adjustbox}{width=\textwidth}
  \begin{tabular}{llll}
    \toprule
    \textbf{Model} & \makecell[l]{\textbf{Number of}\\\textbf{Parameters (M)}} & \textbf{Pre-training Dataset} & \textbf{Data Type} \\
    \midrule
    \multicolumn{4}{c}{\textbf{Protein Language Models}} \\
    \midrule
    ProtBert~\cite{elnaggar2021prottrans} & 420M & UniRef100~\cite{suzek2007uniref} (216M) & General Protein \\
    ESM-2 (150M)~\cite{lin2023evolutionary} & 150M & UniRef50~\cite{suzek2007uniref} (65M) & General Protein \\
    ESM-2 (650M)~\cite{lin2023evolutionary} & 650M & UniRef50~\cite{suzek2007uniref} (65M) & General Protein \\
    \midrule
    \multicolumn{4}{c}{\textbf{Antibody Language Models}} \\
    \midrule
    AbLang-H~\cite{olsen2022ablang} & 86M & OAS~\cite{kovaltsuk2018observed} (14M heavy) & Heavy Chain \\
    AbLang-L~\cite{olsen2022ablang} & 86M & OAS~\cite{kovaltsuk2018observed} (0.24M light) & Light Chain \\
    AntiBERTy~\cite{ruffolo2021deciphering} & 26M & OAS~\cite{kovaltsuk2018observed} (588M) & Heavy, Light Antibody \\
    AntiBERTa2~\cite{barton2024enhancing} & 203M & OAS~\cite{kovaltsuk2018observed} + proprietary (824M) & Heavy, Light Chain \\
    AntiBERTa2-CSSP~\cite{barton2024enhancing} & 202M & OAS~\cite{kovaltsuk2018observed} + structure contrastive (824M) & Heavy, Light Chain \\
    IgBert~\cite{kenlay2024large} & 420M & OAS~\cite{kovaltsuk2018observed} + 2B unpaired + 2M paired & Heavy, Light Chain \\
    \midrule
    \multicolumn{4}{c}{\textbf{Nanobody Language Models}} \\
    \midrule
    NanoBERT~\cite{hadsund2024nanobert} & 15M & INDI~\cite{deszynski2022indi} (10M) & Heavy Nanobody \\
    VHHBERT~\cite{tsuruta2024sars} & 86M & VHHCorpus~\cite{tsuruta2024sars} (2M) & Heavy Nanobody \\
    \bottomrule
  \end{tabular}
  \normalsize
  \end{adjustbox}
\end{table}

\section{Results}
\subsection{Task Pipeline}
NbBench tasks fall into two main categories: sequence-level and token-level prediction. Each category includes two variants, depending on whether the task involves only nanobody sequences or both nanobody and antigen sequences. Antigen embeddings are consistently extracted using ESM-2 (650M), while nanobody sequences are encoded using each respective pretrained model under evaluation.

\textbf{Sequence-Level Prediction.} We consider two types of sequence-level tasks:

\textit{Nanobody-only Tasks:} The model takes the nanobody sequence as input and uses the \texttt{[CLS]} token representation to generate a fixed-length embedding. This embedding is passed through a multilayer perceptron (MLP) for sequence-level classification or regression.
\textit{Nanobody-Antigen Tasks:} For tasks such as antigen-nanobody binding prediction, we first extract a global embedding of the antigen using ESM-2 (650M). This embedding is then concatenated with the \texttt{[CLS]} token from the nanobody sequence. The concatenated representation is pooled and processed through an MLP to generate the final prediction.

\textbf{Token-Level Prediction.} Token-level tasks also fall into two types:

\textit{Nanobody-only Tasks:} The model produces token-level embeddings for each amino acid in the nanobody sequence. These embeddings are independently passed through an MLP to make token-wise predictions (e.g., for CDR classification and CDR infilling).

\textit{Nanobody-Antigen Tasks:} In paratope prediction, we compute antigen embeddings using ESM-2 (650M), then concatenate the antigen representation with each position in the nanobody sequence. The resulting context-aware embeddings are mean pooled across antigen tokens and passed through multiple residual blocks to produce per-residue predictions on the nanobody.

\subsection{Benchmark Results}

We evaluated eleven frozen backbone models—including general protein language models, antibody models, and nanobody-specific models—on the twelve tasks in NbBench. The main evaluation metric for each task is shown in Table~\ref{tab:nbbench-main}, while additional metrics such as precision, recall, and RMSE are provided in Appendix~\ref{subsec:full-results}. All experiments were run using three different random seeds, and we report the average performance with standard deviations. 

\textbf{Easy structure--type tasks.}
Variable\-region classification (VRCls) and nanobody type classification (NbType) are essentially solved: every model exceeds 98\% accuracy.  
Primary--sequence motifs already suffice to separate frameworks from CDRs and to tell VHH from VNAR, so these two tasks are no longer discriminative.

\textbf{Antigen--related tasks favour antibody LMs.}
On \textbf{SARS\mbox{-}2} and \textbf{hIL6} binding as well as \textbf{Paratope} prediction, antibody LMs dominate; \emph{AntiBERTa2\mbox{-}CSSP} reaches 0.92~AUROC for hIL6 and 0.94 for paratope.  
The large immunoglobulin corpora used during pre\-training clearly encode cross\-molecule interaction cues that general protein LMs---and the smaller nanobody LMs---do not capture.


\textbf{Regression remains hard.}
For numerical traits such as \textbf{Thermo} stability and \textbf{Affinity}, all models perform modestly (best Spearman~$\rho\approx0.59$ for stability and $<0.20$ for affinity).  
Pure sequence embeddings---kept frozen here---appear insufficient to recover fine\-grained physicochemical signals without structural or experimental features.

\textbf{No universal winner.}
Antibody LMs give the best overall balance for classification and binding; protein LMs lead on PolyRx.  
These results confirm that model performance is highly task\-specific and that future work should combine domain\-matched pre\-training with structure\-aware or parameter\-efficient fine tuning to push the remaining difficult fronts.

\begin{table*}[htbp]
  \centering
  \caption{
    Performance of various protein, antibody, and nanobody language models on the NbBench benchmark.  
    Each value corresponds to the task's main evaluation metric (e.g., Accuracy, AUROC, Spearman).  
    Shading indicates top-3 performance for each task: darker is better.  
    BR stands for BLOSUM62 Recovery, a soft recovery metric computed using BLOSUM62 substitution matrix. Mean (std) is reported for each experiment using three different seeds.
  }
  \label{tab:nbbench-main}
  \resizebox{\textwidth}{!}{
  \begin{tabular}{lccccccccccc}
    \toprule
    Model & VRCls & CDRInf & SARS-2 & hIL6 & Paratope & Thermo-seq & Thermo-tm & PolyRx & NbType & Affinity-seq & Affinity-score \\
    Metric & Accuracy & BR & AUROC & AUROC & AUROC & Spearman & Spearman & AUROC & Accuracy & Spearman & Spearman \\
    \midrule
    \textbf{Protein Language Models} \\
    \midrule
    ProtBert        & \cellcolor{gray!30}0.9983 (0.0004) & \cellcolor{gray!10}1.520 (0.0039) & \cellcolor{gray!10}0.852 (0.004) & 0.878 (0.011) & 0.916 (0.004) & 0.040 (0.001) & 0.128 (0.002) & \cellcolor{gray!30}0.837 (0.002) & \cellcolor{gray!10}0.957 (0.001) & 0.163 (0.008) & 0.084 (0.015) \\
    ESM-2 (150M)    & \cellcolor{gray!50}0.9989 (0.0004) & 1.499 (0.0034) & 0.834 (0.010) & 0.850 (0.011) & 0.922 (0.003) & 0.389 (0.000) & 0.301 (0.003) & 0.833 (0.004) & 0.994 (0.001) & 0.170 (0.000) & 0.063 (0.000) \\
    ESM-2 (650M)    & \cellcolor{gray!30}0.9984 (0.0003) & \cellcolor{gray!50}1.551 (0.0029) & 0.850 (0.002) & 0.857 (0.005) & 0.928 (0.000) & 0.375 (0.010) & 0.315 (0.000) & \cellcolor{gray!50}0.842 (0.004) & \cellcolor{gray!10}0.995 (0.000) & \cellcolor{gray!10}0.183 (0.003) & 0.089 (0.001) \\
    \midrule
    \textbf{Antibody Language Models} \\
    \midrule
    AbLang-H        & 0.9981 (0.0001) & 1.452 (0.0290) & \cellcolor{gray!50}0.884 (0.000) & \cellcolor{gray!50}0.925 (0.000) & \cellcolor{gray!10}0.933 (0.000) & 0.566 (0.005) & \cellcolor{gray!10}0.533 (0.004) & 0.831 (0.003) & \cellcolor{gray!30}0.997 (0.000) & 0.151 (0.000) & 0.106 (0.000) \\
    AbLang-L        & 0.9984 (0.0002) & 1.360 (0.0144) & 0.811 (0.018) & \cellcolor{gray!10}0.914 (0.011) & \cellcolor{gray!10}0.933 (0.000) & \cellcolor{gray!10}0.585 (0.001) & 0.502 (0.002) & 0.819 (0.002) & 0.988 (0.000) & 0.173 (0.010) & \cellcolor{gray!30}0.128 (0.004) \\
    AntiBERTy       & \cellcolor{gray!30}0.9983 (0.0004) & 1.499 (0.0184) & 0.810 (0.004) & 0.871 (0.003) & \cellcolor{gray!30}0.932 (0.002) & 0.474 (0.006) & 0.385 (0.007) & 0.828 (0.001) & \cellcolor{gray!30}0.999 (0.001) & 0.167 (0.017) & 0.119 (0.013) \\
    AntiBERTa2      & \cellcolor{gray!50}0.9989 (0.0005) & 1.489 (0.0020) & 0.824 (0.001) & 0.887 (0.006) & 0.930 (0.001) & \cellcolor{gray!50}0.587 (0.002) & \cellcolor{gray!30}0.585 (0.003) & \cellcolor{gray!10}0.833 (0.003) & 0.995 (0.000) & \cellcolor{gray!30}0.162 (0.025) & 0.088 (0.001) \\
    AntiBERTa2-CSSP & \cellcolor{gray!30}0.9983 (0.0004) & 1.484 (0.0041) & 0.845 (0.011) & \cellcolor{gray!30}0.916 (0.011) & \cellcolor{gray!50}0.939 (0.013) & \cellcolor{gray!30}0.575 (0.010) & \cellcolor{gray!50}0.593 (0.004) & 0.830 (0.001) & \cellcolor{gray!50}0.998 (0.000) & 0.149 (0.003) & 0.085 (0.011) \\
    IgBert          & \cellcolor{gray!10}0.9981 (0.0003) & 1.311 (0.3290) & 0.817 (0.003) & 0.821 (0.003) & 0.921 (0.001) & 0.351 (0.178) & 0.344 (0.176) & 0.829 (0.010) & 0.993 (0.008) & \cellcolor{gray!50}0.184 (0.006) & \cellcolor{gray!50}0.090 (0.037) \\
    \midrule
    \textbf{Nanobody Language Models} \\
    \midrule
    NanoBERT        & 0.9947 (0.0004) & \cellcolor{gray!30}1.524 (0.0115) & \cellcolor{gray!30}0.876 (0.017) & 0.904 (0.015) & 0.928 (0.001) & 0.440 (0.006) & 0.298 (0.003) & 0.815 (0.004) & 0.988 (0.001) & 0.061 (0.000) & 0.118 (0.000) \\
    VHHBERT         & 0.9904 (0.0004) & 1.244 (0.0219) & 0.800 (0.010) & 0.873 (0.000) & 0.921 (0.002) & 0.484 (0.007) & 0.532 (0.000) & 0.818 (0.003) & 0.986 (0.002) & 0.145 (0.015) & \cellcolor{gray!10}0.115 (0.006) \\
    \bottomrule
  \end{tabular}
  }
\end{table*}

\section{Conclusions and Limitations}
In this work, we introduce NbBench, the first comprehensive benchmark suite for nanobody modeling. It spans eight biologically meaningful tasks—grouped into structure understanding, antigen binding, and biophysical property prediction—and covers nine rigorously curated datasets. To ensure fair and interpretable evaluation, we systematically assess eleven pretrained language models from the protein, antibody, and nanobody domains.

All models are evaluated in a frozen setting, where pretrained parameters remain fixed and only a lightweight classifier is trained on top. This design isolates the representational power of each model and avoids confounding factors from full fine-tuning, allowing us to fairly compare performance across diverse architectures and pretraining domains.

Our results highlight clear performance differences across model types. Antibody-specific models generally outperform others on antigen-related tasks, while nanobody-specific models show strength in thermostability and binding-related predictions. However, no single model dominates across all tasks, revealing that nanobody modeling remains a multifaceted challenge requiring tailored approaches.

NbBench provides a standardized and reproducible foundation for benchmarking nanobody models, but several limitations remain. First, we do not include large-scale LLMs such as GPT-4 or protein-specific foundation models requiring extensive resources, due to accessibility and reproducibility concerns. Second, the benchmark currently focuses on sequence-based tasks, given the limited availability of high-quality nanobody structural data. As more structures become available, future versions of NbBench could incorporate 3D structure-aware tasks such as docking, folding accuracy, or epitope prediction.

We hope NbBench catalyzes progress in nanobody representation learning, design, and downstream applications, and serves as a practical resource for both model development and rigorous evaluation in this growing domain.

\bibliographystyle{unsrt}
\bibliography{references} 

\appendix

\section{Appendix}

\subsection{Experimental Settings for Tasks}
All tasks in NbBench are trained using a shared set of hyperparameters and training settings, as summarized in Table~\ref{tab:training-settings}. All experiments were performed on a single NVIDIA A100 GPU with 80GB of memory.

\begin{table}[htbp]
\centering
\caption{Training configuration for most of the benchmark tasks.}
\label{tab:training-settings}
\begin{tabular}{l|l}
\toprule
\textbf{Config} & \textbf{Value} \\
\midrule
optimizer & AdamW \\
optimizer epsilon & 1e-8 \\
optimizer momentum & $\beta_1=0.9$, $\beta_2=0.999$ \\
weight decay & 0.01 \\
learning rate schedule & linear decay \\
learning rate range & [1e-5, 5e-3] \\
warmup steps & 50 \\
epochs & 50 \\
batch size & 32 \\
gradient accumulation steps & 2 \\
dtype & float32 \\
\bottomrule
\end{tabular}
\end{table}


\subsection{Data Preprocessing}
We curated and preprocessed datasets for each nanobody-related task in NbBench to ensure reliable evaluation and reduce information leakage between training and testing splits. 

\paragraph{Sequence Deduplication and Clustering.}
To avoid data leakage, we removed 100\% identical sequences between training and testing sets. We then applied MMseqs2 clustering with a minimum sequence identity threshold of 0.7 to split sequences into train, validation, and test sets. This ensured that sequences within the same cluster remained in the same split, minimizing cross-split similarity.

\paragraph{Variable Regions Classification and CDRs Infilling.}
Training data were sourced from AbGenBank, manual curation, and structural databases, while testing sequences were extracted from patents. Each residue in the nanobody sequence was labeled according to its region (FR: 0, CDR1: 1, CDR2: 2, CDR3: 3), enabling token-level classification. For the CDRs infilling task, residues in the CDR1, CDR2, and CDR3 regions were masked by replacing them with hyphens ('-'), and language models were tasked with inferring the masked amino acids at these positions. This setup evaluates the models' ability to reconstruct biologically meaningful variable regions based on the surrounding framework sequences.

\paragraph{Antigen-Nanobody Binding.}
We collected binding pairs for SARS-CoV-2 and hIL6. For SARS-CoV-2, the training set includes wild type and variants with mild mutations, while the test set comprises Alpha, Beta, Delta, and Omicron variants to assess generalization. His-tags were removed from antigens such as hIL6 and hTNFa to ensure clean sequence input. Antigen embeddings were precomputed using ESM-2 (650M).

\paragraph{Epitope and Paratope Prediction.}
Structural data from SAbDab-Nano were filtered to include only complexes with resolution < 3.0 Å. Residue pairs within 5 Å were labeled as interacting. Token-level binary classification was applied to the nanobody (paratope) sequences.

\paragraph{Stability and Polyreactivity.}
For nanobody thermostability prediction, we compiled 764 sequences from multiple databases and stratified them by melting temperature (TM). The average sequence similarity between train and test sets was around 75\%. For polyreactivity, high and low reactivity classes were separately clustered and merged afterward to ensure balanced representation and limited inter-class similarity.

\paragraph{Nanobody Type and Affinity prediction.}
We classified antibodies into five types: VHH, VH, VNAR, Vkappa, and Vlambda, using balanced splits with inter-set similarity controlled (e.g., 76–80\%). For affinity prediction, only sequences with binding scores $\geq 5$ or $\leq 1$ were retained for binary classification.

\paragraph{Similarity Control.}
Across all tasks, MMseqs2's "easy-search" module was used to compute the average similarity between training and testing sequences. All tasks reported train-test similarities between 73\% and 84\%, ensuring non-trivial generalization challenges.

\subsection{Details for Antigen Binding and Paratope Prediction Tasks}
In the antigen binding prediction task, the model input consists of a nanobody sequence and a precomputed antigen embedding. The nanobody sequence is encoded using a pretrained nanobody model, and the \texttt{[CLS]} token embedding is extracted from the final layer. Separately, the antigen embedding is generated using ESM-2 (650M) and mean-pooled along the sequence dimension. The pooled antigen embedding is then concatenated with the nanobody \texttt{[CLS]} embedding. The combined representation is passed through a linear classifier to predict binding outcomes, supporting both binary and multi-class classification, as well as regression settings depending on the task configuration.

For paratope prediction, the model receives the nanobody sequence along with an antigen embedding. The nanobody sequence is processed to produce token-level embeddings for each amino acid. Simultaneously, the antigen embedding is mean-pooled and expanded across the length of the nanobody sequence. At each amino acid position, the token embedding and the pooled antigen embedding are concatenated to form context-aware representations. These representations are further processed through three residual blocks and finally passed through a token-wise classifier to predict paratope labels for each residue.

\subsection{BLOSUM62 Recovery}
In the CDR Infilling task, we additionally report \textbf{BLOSUM62 Recovery (BR)}, which assigns partial credit for biochemically similar substitutions, complementing the strict evaluation of Exact Match (EM).

\vspace{0.5em}
\noindent
\textbf{Motivation and biological significance:}  
The BLOSUM62 matrix reflects empirical substitution probabilities between amino acids observed in evolutionarily related proteins. Many amino acid substitutions are functionally tolerated because they preserve key chemical properties such as charge, hydrophobicity, or size. Therefore, even if a model's prediction does not exactly match the ground-truth residue, it may still maintain biological function if the substitution is favorable according to BLOSUM62.  
BR captures this notion of functional robustness, providing a more biologically meaningful evaluation of sequence recovery, especially for applications like nanobody design, where partial recovery can still yield viable binding or stability.

\vspace{0.5em}
\noindent
\textbf{Calculation:}  
Given the ground-truth sequence $\mathbf{y}$ and the predicted sequence $\hat{\mathbf{y}}$, the BLOSUM62 score for each position is retrieved using the substitution matrix. The overall BLOSUM62 Recovery score is computed as the average substitution score across all valid positions:
\[
\text{BR} = \frac{1}{N} \sum_{i=1}^{N} \text{BLOSUM62}(y_i, \hat{y}_i)
\]
where $N$ denotes the number of valid amino acid positions (excluding padding tokens). In practice, scores are normalized if needed for comparability.

Furthermore, we separately compute token-level BR across the entire sequence and CDR-specific BR scores (for CDR1, CDR2, and CDR3), allowing finer-grained evaluation.

The pseudocode used to calculate BR is summarized below, and the full BLOSUM62 substitution matrix used for scoring is provided in Table~\ref{tab:full-blosum62}.

\begin{itemize}
    \item For each sample, identify valid amino acid positions.
    \item Partition the positions into CDR1, CDR2, and CDR3.
    \item For each position, retrieve the BLOSUM62 score between the ground-truth and predicted residues.
    \item Accumulate scores separately for the full sequence and each CDR region.
    \item Report the mean BLOSUM62 score overall and per CDR.
\end{itemize}

\vspace{0.5em}
\noindent
\textbf{Comparison with Exact Match (EM):}  
While EM measures the fraction of residues that are predicted perfectly, it does not distinguish between near-correct and completely incorrect predictions. In contrast, BR provides a graded measure that rewards chemically plausible substitutions, thus offering a more nuanced assessment of model outputs.

\begin{table}[htbp]
\centering
\caption{Complete BLOSUM62 substitution matrix.}
\label{tab:full-blosum62}
\begin{adjustbox}{width=\textwidth}
\begin{tabular}{c|rrrrrrrrrrrrrrrrrrrrrrrr}
\toprule
 & A & R & N & D & C & Q & E & G & H & I & L & K & M & F & P & S & T & W & Y & V & B & Z & X & * \\
\midrule
A &  4 & -1 & -2 & -2 &  0 & -1 & -1 &  0 & -2 & -1 & -1 & -1 & -1 & -2 & -1 &  1 &  0 & -3 & -2 &  0 & -2 & -1 &  0 & -4 \\
R & -1 &  5 &  0 & -2 & -3 &  1 &  0 & -2 &  0 & -3 & -2 &  2 & -1 & -3 & -2 & -1 & -1 & -3 & -2 & -3 & -1 &  0 & -1 & -4 \\
N & -2 &  0 &  6 &  1 & -3 &  0 &  0 &  0 &  1 & -3 & -3 &  0 & -2 & -3 & -2 &  1 &  0 & -4 & -2 & -3 &  3 &  0 & -1 & -4 \\
D & -2 & -2 &  1 &  6 & -3 &  0 &  2 & -1 & -1 & -3 & -4 & -1 & -3 & -3 & -1 &  0 & -1 & -4 & -3 & -3 &  4 &  1 & -1 & -4 \\
C &  0 & -3 & -3 & -3 &  9 & -3 & -4 & -3 & -3 & -1 & -1 & -3 & -1 & -2 & -3 & -1 & -1 & -2 & -2 & -1 & -3 & -3 & -2 & -4 \\
Q & -1 &  1 &  0 &  0 & -3 &  5 &  2 & -2 &  0 & -3 & -2 &  1 &  0 & -3 & -1 &  0 & -1 & -2 & -1 & -2 &  0 &  3 & -1 & -4 \\
E & -1 &  0 &  0 &  2 & -4 &  2 &  5 & -2 &  0 & -3 & -3 &  1 & -2 & -3 & -1 &  0 & -1 & -3 & -2 & -2 &  1 &  4 & -1 & -4 \\
G &  0 & -2 &  0 & -1 & -3 & -2 & -2 &  6 & -2 & -4 & -4 & -2 & -3 & -3 & -2 &  0 & -2 & -2 & -3 & -3 & -1 & -2 & -1 & -4 \\
H & -2 &  0 &  1 & -1 & -3 &  0 &  0 & -2 &  8 & -3 & -3 & -1 & -2 & -1 & -2 & -1 & -2 & -2 &  2 & -3 &  0 &  0 & -1 & -4 \\
I & -1 & -3 & -3 & -3 & -1 & -3 & -3 & -4 & -3 &  4 &  2 & -3 &  1 &  0 & -3 & -2 & -1 & -3 & -1 &  3 & -3 & -3 & -1 & -4 \\
L & -1 & -2 & -3 & -4 & -1 & -2 & -3 & -4 & -3 &  2 &  4 & -2 &  2 &  0 & -3 & -2 & -1 & -2 & -1 &  1 & -4 & -3 & -1 & -4 \\
K & -1 &  2 &  0 & -1 & -3 &  1 &  1 & -2 & -1 & -3 & -2 &  5 & -1 & -3 & -1 &  0 & -1 & -3 & -2 & -2 &  0 &  1 & -1 & -4 \\
M & -1 & -1 & -2 & -3 & -1 &  0 & -2 & -3 & -2 &  1 &  2 & -1 &  5 &  0 & -2 & -1 & -1 & -1 & -1 &  1 & -3 & -1 & -1 & -4 \\
F & -2 & -3 & -3 & -3 & -2 & -3 & -3 & -3 & -1 &  0 &  0 & -3 &  0 &  6 & -4 & -2 & -2 &  1 &  3 & -1 & -3 & -3 & -1 & -4 \\
P & -1 & -2 & -2 & -1 & -3 & -1 & -1 & -2 & -2 & -3 & -3 & -1 & -2 & -4 &  7 & -1 & -1 & -4 & -3 & -2 & -2 & -1 & -2 & -4 \\
S &  1 & -1 &  1 &  0 & -1 &  0 &  0 &  0 & -1 & -2 & -2 &  0 & -1 & -2 & -1 &  4 &  1 & -3 & -2 & -2 &  0 &  0 &  0 & -4 \\
T &  0 & -1 &  0 & -1 & -1 & -1 & -1 & -2 & -2 & -1 & -1 & -1 & -1 & -2 & -1 &  1 &  5 & -2 & -2 &  0 & -1 &  0 & -4 & -4 \\
W & -3 & -3 & -4 & -4 & -2 & -2 & -3 & -2 & -2 & -3 & -2 & -3 & -1 &  1 & -4 & -3 & -2 & 11 &  2 & -3 & -4 & -3 & -2 & -4 \\
Y & -2 & -2 & -2 & -3 & -2 & -1 & -2 & -3 &  2 & -1 & -1 & -2 & -1 &  3 & -3 & -2 & -2 &  2 &  7 & -1 & -3 & -2 & -1 & -4 \\
V &  0 & -3 & -3 & -3 & -1 & -2 & -2 & -3 & -3 &  3 &  1 & -2 &  1 & -1 & -2 & -2 &  0 & -3 & -1 &  4 & -3 & -2 & -1 & -4 \\
B & -2 & -1 &  3 &  4 & -3 &  0 &  1 & -1 &  0 & -3 & -4 &  0 & -3 & -3 & -2 & -2 &  0 & -4 & -3 & -3 &  4 &  1 & -1 & -4 \\
Z & -1 &  0 &  0 &  1 & -3 &  3 &  4 & -2 &  0 & -3 & -3 &  1 & -1 & -3 & -1 &  0 & -1 & -3 & -2 & -2 &  1 &  4 & -1 & -4 \\
X &  0 & -1 & -1 & -1 & -2 & -1 & -1 & -1 & -1 & -1 & -1 & -1 & -1 & -1 & -2 &  0 &  0 & -2 & -1 & -1 & -1 & -1 & -1 & -4 \\
* & -4 & -4 & -4 & -4 & -4 & -4 & -4 & -4 & -4 & -4 & -4 & -4 & -4 & -4 & -4 & -4 & -4 & -4 & -4 & -4 & -4 & -4 & -4 &  1 \\
\bottomrule
\end{tabular}
\end{adjustbox}
\end{table}

\subsection{Full Results}
\label{subsec:full-results}

In this section, we report the full evaluation results of each task in NbBench, grouped by task category. For each task, we provide a detailed performance table covering all evaluated models.

\vspace{0.5em}
\noindent
\textbf{Structure and Region Understanding.}  
We evaluate two tasks in this category: Variable Region Classification (VRCls) and CDR Infilling (CDRInf).  
Table~\ref{tab:vrcls} summarizes the performance on VRCls using standard classification metrics such as accuracy, precision, recall, and F1 score.  
For the CDRInf task, we evaluate both exact match (EM) and BLOSUM62 Recovery (BR) scores for each of the three CDRs and their aggregate, as shown in Table~\ref{tab:cdrinf}.  
These results allow us to assess both strict correctness and biochemical plausibility in recovered sequences.

\vspace{0.5em}
\noindent
\textbf{Binding Prediction.}  
We report model performance on three classification tasks: binding to SARS-CoV-2, binding to human IL-6 (hIL6), and paratope prediction.  
Each task is evaluated using AUROC, AUPRC, accuracy, precision, recall, and F1 score.  
Table~\ref{tab:binding-metrics} provides a comprehensive overview of results across these tasks.  
These metrics help evaluate how well models capture structural determinants of binding and functional interfaces.

\vspace{0.5em}
\noindent
\textbf{Thermostability and Affinity Prediction.}  
We assess model performance on four regression tasks: Thermo-seq and Thermo-tm (predicting thermal stability), Affinity-seq and Affinity-score (predicting binding affinity).  
Each task is evaluated using the Spearman correlation, $R^2$, RMSE, and MAE. 
Results are shown in the second half of Table~\ref{tab:regression-metrics}, which has been reused for consistency.  
These evaluations capture how well models generalize to scalar-valued biophysical properties.

\vspace{0.5em}
\noindent
\textbf{Polyreactivity Prediction.}  
This classification task evaluates whether a nanobody exhibits nonspecific binding across diverse targets.  
As shown in Table~\ref{tab:polyreactivity-metrics}, we report standard classification metrics (AUROC, AUPRC, accuracy, precision, recall, F1).  
Polyreactivity prediction is essential for screening nanobody candidates for developability.

\vspace{0.5em}
\noindent
\textbf{Nanobody Type Prediction.}  
We evaluate model accuracy in classifying the nanobody’s type (e.g., VHH from alpaca vs. VNAR from shark).  
Results in Table~\ref{tab:nanobody-type-metrics} show near-ceiling performance for several models, with high precision and recall across the board.

\begin{table}[htbp]
\centering
\caption{Performance on the VRCls task (normalized to [0, 1]).}
\label{tab:vrcls}
\begin{tabular}{l|cccc}
\toprule
Model & Accuracy & Precision & Recall & F1 Score \\
\midrule
ProtBert & 0.9983 (0.0004) & 0.9974 (0.0001) & 0.9975 (0.0002) & 0.9975 (0.0002) \\
ESM-2 (150M) & \textbf{0.9989 (0.0004)} & 0.9982 (0.0004) & 0.9987 (0.0005) & \textbf{0.9985 (0.0001)} \\
ESM-2 (650M) & 0.9984 (0.0003) & 0.9979 (0.0001) & \textbf{0.9990 (0.0002)} & 0.9984 (0.0003) \\
AbLang-H & 0.9981 (0.0001) & 0.9964 (0.0002) & 0.9959 (0.0004) & 0.9962 (0.0003) \\
AbLang-L & 0.9984 (0.0002) & 0.9941 (0.0003) & 0.9915 (0.0004) & 0.9928 (0.0001) \\
AntiBERTy & 0.9983 (0.0004) & 0.9967 (0.0004) & 0.9965 (0.0002) & 0.9966 (0.0002) \\
AntiBERTa2 & 0.9989 (0.0005) & \textbf{0.9986 (0.0002)} & 0.9972 (0.0001) & 0.9979 (0.0001) \\
AntiBERTa2-CSSP & 0.9983 (0.0004) & 0.9968 (0.0003) & 0.9977 (0.0004) & 0.9972 (0.0004) \\
IgBert & 0.9981 (0.0003) & 0.9980 (0.0005) & 0.9949 (0.0003) & 0.9965 (0.0003) \\
NanoBERT & 0.9947 (0.0004) & 0.9916 (0.0003) & 0.9885 (0.0004) & 0.9900 (0.0003) \\
VHHBERT & 0.9904 (0.0004) & 0.9836 (0.0001) & 0.9776 (0.0002) & 0.9806 (0.0002) \\
\bottomrule
\end{tabular}
\end{table}

\vspace{1em}

\begin{table}[htbp]
\centering
\caption{Performance on the CDRInf task. EM and BR stand for exact match and BLOSUM62 Recovery, respectively.}
\label{tab:cdrinf}
\begin{adjustbox}{width=\textwidth}
\begin{tabular}{l|cccccccc}
\toprule
Model & CDR1 EM & CDR2 EM & CDR3 EM & CDRs EM & CDR1 BR & CDR2 BR & CDR3 BR & CDRs BR\\
\midrule
ProtBert & 0.543 (0.0008) & 0.533 (0.0002) & \textbf{0.271 (0.0005)} & 0.406 (0.0004) & 2.597 (0.0056) & 2.403 (0.0021) & \textbf{0.537 (0.0068)} & 1.525 (0.0039) \\
ESM-2 (150M) & 0.543 (0.0005) & 0.531 (0.0006) & 0.265 (0.0005) & 0.402 (0.0003) & 2.576 (0.0049) & 2.400 (0.0024) & 0.499 (0.0055) & 1.500 (0.0034) \\
ESM-2 (650M) & \textbf{0.553 (0.0005)} & \textbf{0.539 (0.0003)} & 0.269 (0.0003) & \textbf{0.408 (0.0002)} & \textbf{2.651 (0.0010)} & \textbf{2.464 (0.0021)} & 0.536 (0.0054) & \textbf{1.553 (0.0029)} \\
AbLang-H & 0.533 (0.0038) & 0.524 (0.0013) & 0.263 (0.0018) & 0.397 (0.0020) & 2.552 (0.0117) & 2.340 (0.0108) & 0.442 (0.0482) & 1.452 (0.0290) \\
AbLang-L & 0.520 (0.0009) & 0.516 (0.0006) & 0.249 (0.0025) & 0.385 (0.0011) & 2.441 (0.0258) & 2.292 (0.0073) & 0.336 (0.0379) & 1.360 (0.0144) \\
AntiBERTy & 0.541 (0.0014) & 0.529 (0.0013) & 0.266 (0.0020) & 0.401 (0.0013) & 2.560 (0.0088) & 2.382 (0.0077) & 0.514 (0.0335) & 1.499 (0.0184) \\
AntiBERTa2 & 0.538 (0.0006) & 0.531 (0.0002) & 0.263 (0.0006) & 0.399 (0.0003) & 2.539 (0.0043) & 2.405 (0.0010) & 0.494 (0.0021) & 1.489 (0.0020) \\
AntiBERTa2-CSSP & 0.535 (0.0010) & 0.528 (0.0015) & 0.263 (0.0008) & 0.398 (0.0010) & 2.515 (0.0059) & 2.385 (0.0118) & 0.506 (0.0041) & 1.484 (0.0041) \\
IgBert & 0.516 (0.0491) & 0.502 (0.0526) & 0.249 (0.0268) & 0.380 (0.0389) & 2.383 (0.3623) & 2.190 (0.3770) & 0.321 (0.2880) & 1.311 (0.3290) \\
NanoBERT & 0.546 (0.0016) & 0.530 (0.0030) & 0.266 (0.0015) & 0.403 (0.0011) & 2.637 (0.0078) & 2.402 (0.0210) & 0.515 (0.0182) & 1.524 (0.0115) \\
VHHBERT & 0.464 (0.0037) & 0.490 (0.0014) & 0.256 (0.0005) & 0.367 (0.0007) & 1.983 (0.0394) & 2.101 (0.0104) & 0.436 (0.0326) & 1.244 (0.0219) \\
\bottomrule
\end{tabular}
\end{adjustbox}
\end{table}

\begin{table*}[htbp]
\centering
\caption{Performance of various models across three tasks: SARS-CoV-2 Binding Prediction, hIL6 Binding Prediction, and Paratope Prediction. Each metric reports the mean and standard deviation over three seeds.}
\label{tab:binding-paratope}
\begin{adjustbox}{width=\textwidth}
\begin{tabular}{llcccccc}
\toprule
\textbf{Task} & \textbf{Model} & \textbf{AUROC} & \textbf{AUPRC} & \textbf{Acc} & \textbf{Prec} & \textbf{Rec} & \textbf{F1} \\
\midrule
\multirow{11}{*}{SARS-CoV-2}
& ProtBert & 0.852 (0.004) & 0.785 (0.008) & 0.847 (0.005) & 0.845 (0.031) & 0.565 (0.020) & 0.676 (0.011) \\
& ESM-2 (150M) & 0.834 (0.010) & 0.758 (0.012) & 0.831 (0.005) & 0.823 (0.023) & 0.519 (0.021) & 0.636 (0.015) \\
& ESM-2 (650M) & 0.850 (0.002) & 0.793 (0.003) & 0.847 (0.003) & 0.808 (0.037) & 0.606 (0.033) & 0.692 (0.008) \\
& AbLang-H & \textbf{0.884} (0.000) & \textbf{0.850} (0.000) & \textbf{0.889} (0.000) & \textbf{0.927} (0.000) & \textbf{0.660} (0.000) & \textbf{0.771} (0.000) \\
& AbLang-L & 0.811 (0.018) & 0.737 (0.033) & 0.824 (0.013) & 0.868 (0.052) & 0.448 (0.019) & 0.591 (0.029) \\
& AntiBERTy & 0.810 (0.004) & 0.719 (0.007) & 0.818 (0.004) & 0.852 (0.004) & 0.436 (0.035) & 0.576 (0.022) \\
& AntiBERTa2 & 0.824 (0.001) & 0.760 (0.001) & 0.834 (0.000) & 0.769 (0.001) & 0.596 (0.000) & 0.672 (0.000) \\
& AntiBERTa2-CSSP & 0.845 (0.011) & 0.793 (0.013) & 0.850 (0.011) & 0.831 (0.021) & 0.591 (0.043) & 0.690 (0.028) \\
& IgBert & 0.817 (0.003) & 0.743 (0.008) & 0.817 (0.014) & 0.909 (0.045) & 0.399 (0.082) & 0.549 (0.068) \\
& NanoBERT & 0.876 (0.017) & 0.829 (0.029) & 0.874 (0.019) & 0.873 (0.039) & 0.650 (0.039) & 0.745 (0.040) \\
& VHHBERT & 0.800 (0.010) & 0.712 (0.007) & 0.811 (0.007) & 0.811 (0.082) & 0.454 (0.101) & 0.573 (0.061) \\
\midrule
\multirow{11}{*}{hIL6}
& ProtBert & 0.878 (0.011) & 0.578 (0.031) & 0.973 (0.002) & 0.888 (0.052) & 0.373 (0.076) & 0.519 (0.064) \\
& ESM-2 (150M) & 0.850 (0.011) & 0.586 (0.026) & 0.975 (0.001) & 0.948 (0.001) & 0.380 (0.022) & 0.540 (0.023) \\
& ESM-2 (650M) & 0.857 (0.005) & 0.550 (0.011) & 0.964 (0.002) & 0.971 (0.003) & 0.088 (0.053) & 0.159 (0.086) \\
& AbLang-H & 0.925 (0.000) & 0.737 (0.000) & 0.979 (0.000) & 0.937 (0.000) & 0.512 (0.000) & 0.662 (0.000) \\
& AbLang-L & 0.914 (0.011) & 0.672 (0.016) & 0.978 (0.002) & 0.944 (0.012) & 0.482 (0.003) & 0.639 (0.002) \\
& AntiBERTy & 0.871 (0.003) & 0.565 (0.009) & 0.977 (0.002) & 0.932 (0.006) & 0.444 (0.010) & 0.602 (0.007) \\
& AntiBERTa2 & 0.887 (0.006) & 0.615 (0.020) & 0.976 (0.001) & 0.926 (0.013) & 0.445 (0.032) & 0.597 (0.025) \\
& AntiBERTa2-CSSP & \textbf{0.916} (0.011) & \textbf{0.725} (0.015) & \textbf{0.981} (0.001) & \textbf{0.939} (0.000) & \textbf{0.548} (0.020) & \textbf{0.696} (0.018) \\
& IgBert & 0.821 (0.003) & 0.430 (0.020) & 0.969 (0.000) & 0.929 (0.022) & 0.247 (0.005) & 0.389 (0.006) \\
& NanoBERT & 0.904 (0.015) & 0.687 (0.021) & 0.978 (0.001) & 0.948 (0.009) & 0.481 (0.007) & 0.638 (0.005) \\
& VHHBERT & 0.873 (0.000) & 0.593 (0.000) & 0.976 (0.000) & 0.955 (0.000) & 0.426 (0.000) & 0.589 (0.000) \\
\midrule
\multirow{11}{*}{Paratope}
& ProtBert & 0.916 (0.004) & 0.693 (0.008) & 0.896 (0.000) & 0.809 (0.001) & 0.817 (0.001) & 0.813 (0.000) \\
& ESM-2 (150M) & 0.922 (0.003) & 0.714 (0.002) & 0.901 (0.001) & 0.820 (0.012) & 0.819 (0.004) & 0.819 (0.003) \\
& ESM-2 (650M) & 0.928 (0.000) & 0.718 (0.000) & 0.900 (0.000) & 0.818 (0.000) & 0.820 (0.000) & 0.819 (0.000) \\
& AbLang-H & 0.933 (0.000) & 0.756 (0.000) & 0.902 (0.000) & 0.821 (0.000) & 0.822 (0.000) & 0.821 (0.000) \\
& AbLang-L & 0.933 (0.000) & 0.742 (0.009) & 0.900 (0.000) & 0.821 (0.001) & 0.811 (0.018) & 0.816 (0.010) \\
& AntiBERTy & 0.932 (0.002) & 0.739 (0.004) & 0.902 (0.001) & 0.821 (0.003) & 0.821 (0.003) & 0.821 (0.002) \\
& AntiBERTa2 & 0.930 (0.001) & 0.746 (0.001) & 0.903 (0.002) & 0.825 (0.001) & 0.816 (0.020) & 0.821 (0.010) \\
& AntiBERTa2-CSSP & \textbf{0.939} (0.013) & \textbf{0.765} (0.004) & \textbf{0.906} (0.002) & \textbf{0.831} (0.001) & \textbf{0.822} (0.001) & \textbf{0.826} (0.001) \\
& IgBert & 0.921 (0.001) & 0.687 (0.002) & 0.896 (0.001) & 0.808 (0.002) & 0.820 (0.006) & 0.814 (0.002) \\
& NanoBERT & 0.928 (0.001) & 0.719 (0.002) & 0.897 (0.001) & 0.812 (0.004) & 0.810 (0.004) & 0.811 (0.001) \\
& VHHBERT & 0.921 (0.002) & 0.695 (0.007) & 0.890 (0.001) & 0.802 (0.002) & 0.794 (0.004) & 0.798 (0.002) \\
\bottomrule
\end{tabular}
\end{adjustbox}
\end{table*}

\begin{table*}[htbp]
\centering
\caption{Performance on regression tasks (Thermo-seq, Thermo-tm, Affinity-seq and Affinity-score). Each task reports Spearman coefficient, $R^2$, RMSE and MAE (mean $\pm$ std).}
\label{tab:regression-metrics}

\begin{adjustbox}{width=\textwidth}
\begin{tabular}{l|cccc|cccc}
\toprule
\multirow{2}{*}{\textbf{Model}} 
& \multicolumn{4}{c|}{\textbf{Thermo-seq}} 
& \multicolumn{4}{c}{\textbf{Thermo-tm}} \\
\cmidrule(lr){2-5} \cmidrule(lr){6-9}
& Spearman & $R^2$ & RMSE & MAE
& Spearman & $R^2$ & RMSE & MAE \\
\midrule
ProtBert & 0.040 (0.001) & 0.003 (0.000) & 20.689 (0.048) & 18.767 (0.052)
& 0.128 (0.002) & 0.013 (0.000) & 24.797 (0.054) & 22.886 (0.107) \\
ESM-2 (150M) & 0.389 (0.000) & 0.273 (0.000) & 9.946 (0.000) & 7.771 (0.000)
& 0.301 (0.003) & 0.113 (0.001) & 9.255 (0.025) & 7.810 (0.025) \\
ESM-2 (650M) & 0.375 (0.010) & 0.154 (0.005) & 10.099 (0.050) & 7.683 (0.026)
& 0.315 (0.000) & 0.096 (0.000) & 9.178 (0.000) & 7.842 (0.000) \\
AbLang-H & 0.566 (0.005) & 0.413 (0.012) & 10.482 (0.227) & 8.239 (0.146)
& 0.533 (0.004) & 0.272 (0.006) & 9.916 (0.110) & 7.936 (0.107) \\
AbLang-L & \textbf{0.585 (0.001)} & 0.412 (0.002) & 9.161 (0.171) & 7.173 (0.118)
& 0.502 (0.002) & 0.264 (0.003) & 8.855 (0.105) & 7.143 (0.054) \\
AntiBERTy & 0.474 (0.006) & 0.345 (0.009) & 9.187 (0.204) & 7.050 (0.119)
& 0.385 (0.007) & 0.140 (0.003) & 9.640 (0.069) & 7.888 (0.063) \\
AntiBERTa2 & 0.587 (0.002) & 0.451 (0.000) & \textbf{8.493 (0.001)} & \textbf{6.533 (0.000)}
& 0.585 (0.003) & 0.343 (0.002) & 8.615 (0.142) & 6.723 (0.145) \\
AntiBERTa2-CSSP & 0.575 (0.010) & \textbf{0.459 (0.010)} & 8.808 (0.151) & 6.708 (0.097)
& \textbf{0.593 (0.004)} & \textbf{0.361 (0.004)} & \textbf{8.469 (0.109)} & \textbf{6.839 (0.083)} \\
IgBert & 0.351 (0.178) & 0.156 (0.098) & 10.028 (0.344) & 7.873 (0.168)
& 0.344 (0.176) & 0.144 (0.111) & 9.109 (0.358) & 7.544 (0.337) \\
NanoBERT & 0.440 (0.006) & 0.241 (0.009) & 9.590 (0.073) & 7.380 (0.057)
& 0.298 (0.003) & 0.053 (0.001) & 10.011 (0.037) & 7.949 (0.013) \\
VHHBERT & 0.484 (0.007) & 0.390 (0.004) & 9.318 (0.053) & 7.540 (0.021)
& 0.532 (0.000) & 0.310 (0.000) & 8.701 (0.000) & 6.883 (0.000) \\
\bottomrule
\end{tabular}
\end{adjustbox}
\vspace{1.2em}

\begin{adjustbox}{width=\textwidth}
\begin{tabular}{l|cccc|cccc}
\toprule
\multirow{2}{*}{\textbf{Model}} 
& \multicolumn{4}{c|}{\textbf{Affinity-seq}} 
& \multicolumn{4}{c}{\textbf{Affinity-score}} \\
\cmidrule(lr){2-5} \cmidrule(lr){6-9}
& Spearman & $R^2$ & RMSE & MAE
& Spearman & $R^2$ & RMSE & MAE \\
\midrule
ProtBert & 0.163 (0.008) & 0.044 (0.006) & 4.003 (0.025) & 3.358 (0.029)
& 0.084 (0.015) & 0.030 (0.008) & 1.787 (0.006) & 1.296 (0.119) \\
ESM-2 (150M) & 0.170 (0.000) & 0.073 (0.000) & 3.935 (0.000) & 3.319 (0.000)
& 0.063 (0.000) & 0.037 (0.000) & 1.842 (0.000) & \textbf{1.066 (0.000)} \\
ESM-2 (650M) & 0.183 (0.003) & \textbf{0.082 (0.001)} & 3.464 (0.006) & \textbf{2.779 (0.003)}
& 0.089 (0.001) & 0.052 (0.006) & 1.855 (0.036) & 1.267 (0.236) \\
AbLang-H & 0.151 (0.000) & 0.025 (0.000) & 5.031 (0.000) & 4.537 (0.000)
& 0.106 (0.000) & 0.063 (0.000) & 1.898 (0.000) & 1.608 (0.000) \\
AbLang-L & 0.173 (0.010) & 0.053 (0.008) & 4.073 (0.722) & 3.316 (0.809)
& \textbf{0.128 (0.004)} & 0.058 (0.007) & 2.082 (0.125) & 1.821 (0.153) \\
AntiBERTy & 0.167 (0.017) & 0.074 (0.017) & 3.532 (0.456) & 3.053 (0.544)
& 0.119 (0.013) & 0.072 (0.003) & 1.915 (0.106) & 1.408 (0.194) \\
AntiBERTa2 & 0.162 (0.025) & 0.041 (0.007) & 3.534 (0.731) & 2.840 (0.844)
& 0.088 (0.001) & 0.053 (0.009) & 2.150 (0.092) & 1.630 (0.409) \\
AntiBERTa2-CSSP & 0.149 (0.003) & 0.067 (0.006) & 3.921 (0.940) & 3.307 (1.052)
& 0.085 (0.011) & 0.066 (0.014) & 2.057 (0.113) & 1.418 (0.450) \\
IgBert & \textbf{0.184 (0.006)} & 0.082 (0.013) & \textbf{3.369 (0.582)} & 2.670 (0.666)
& 0.090 (0.037) & 0.065 (0.001) & 1.908 (0.236) & 1.285 (0.185) \\
NanoBERT & 0.061 (0.000) & 0.006 (0.000) & 3.518 (0.000) & 2.811 (0.000)
& 0.118 (0.000) & 0.022 (0.000) & 1.895 (0.000) & 1.575 (0.000) \\
VHHBERT & 0.145 (0.015) & 0.044 (0.019) & 3.110 (0.152) & 2.248 (0.178)
& 0.115 (0.006) & \textbf{0.070 (0.003)} & \textbf{2.092 (0.130)} & 1.667 (0.402) \\
\bottomrule
\end{tabular}
\end{adjustbox}
\end{table*}


\begin{table}[htbp]
\centering
\caption{Performance on Polyreactivity prediction task. The task reports AUROC, AUPRC, Accuracy (Acc), Precision (Prec), Recall (Rec), and F1.}
\label{tab:polyreactivity-metrics}
\begin{adjustbox}{width=\textwidth}
\begin{tabular}{l|cccccc}
\toprule
\textbf{Model} & \textbf{AUROC} & \textbf{AUPRC} & \textbf{Acc} & \textbf{Prec} & \textbf{Rec} & \textbf{F1} \\
\midrule
ProtBert & 0.837 (0.002) & 0.844 (0.002) & 0.766 (0.002) & \textbf{0.787 (0.001)} & 0.775 (0.003) & 0.781 (0.002) \\
ESM-2 (150M) & 0.833 (0.004) & 0.840 (0.006) & 0.764 (0.003) & 0.784 (0.003) & 0.775 (0.008) & 0.779 (0.004) \\
ESM-2 (650M) & \textbf{0.842 (0.004)} & \textbf{0.847 (0.005)} & \textbf{0.773 (0.004)} & 0.783 (0.003) & 0.798 (0.008) & \textbf{0.791 (0.006)} \\
AbLang-H & 0.831 (0.003) & 0.839 (0.004) & 0.765 (0.003) & 0.767 (0.003) & \textbf{0.805 (0.007)} & 0.785 (0.005) \\
AbLang-L & 0.819 (0.002) & 0.823 (0.002) & 0.749 (0.002) & 0.771 (0.011) & 0.761 (0.033) & 0.765 (0.010) \\
AntiBERTy & 0.828 (0.001) & 0.835 (0.001) & 0.758 (0.002) & 0.771 (0.013) & 0.785 (0.029) & 0.777 (0.008) \\
AntiBERTa2 & 0.833 (0.003) & 0.840 (0.004) & 0.761 (0.004) & 0.778 (0.005) & 0.775 (0.010) & 0.776 (0.007) \\
AntiBERTa2-CSSP & 0.830 (0.001) & 0.838 (0.001) & 0.760 (0.003) & 0.779 (0.010) & 0.775 (0.022) & 0.776 (0.007) \\
IgBert & 0.829 (0.010) & 0.836 (0.011) & 0.762 (0.008) & 0.784 (0.006) & 0.790 (0.016) & 0.786 (0.010) \\
NanoBERT & 0.815 (0.004) & 0.823 (0.005) & 0.747 (0.003) & 0.755 (0.007) & 0.785 (0.012) & 0.769 (0.006) \\
VHHBERT & 0.818 (0.003) & 0.823 (0.002) & 0.750 (0.003) & 0.767 (0.005) & 0.770 (0.007) & 0.768 (0.005) \\
\bottomrule
\end{tabular}
\end{adjustbox}
\end{table}


\begin{table}[htbp]
\centering
\caption{Performance on Nanobody Type prediction task. The task reports Accuracy (Acc), Precision (Prec), Recall (Rec), and F1.}
\label{tab:nanobody-type-metrics}
\begin{adjustbox}{width=\textwidth}
\begin{tabular}{l|cccc}
\toprule
\textbf{Model} & \textbf{Acc} & \textbf{Prec} & \textbf{Rec} & \textbf{F1} \\
\midrule
ProtBert & 0.957 (0.001) & 0.995 (0.002) & 0.997 (0.000) & 0.996 (0.001) \\
ESM-2 (150M) & 0.994 (0.001) & 0.995 (0.000) & 0.988 (0.001) & 0.992 (0.000) \\
ESM-2 (650M) & 0.995 (0.000) & 0.996 (0.000) & 0.994 (0.004) & 0.995 (0.002) \\
AbLang-H & 0.997 (0.000) & 0.998 (0.000) & \textbf{0.997 (0.000)} & \textbf{0.997 (0.000)} \\
AbLang-L & 0.988 (0.000) & 0.990 (0.000) & 0.948 (0.006) & 0.966 (0.003) \\
AntiBERTy & \textbf{0.999 (0.001)} & \textbf{0.999 (0.001)} & 0.996 (0.003) & \textbf{0.997 (0.001)} \\
AntiBERTa2 & 0.995 (0.000) & 0.996 (0.000) & 0.953 (0.009) & 0.972 (0.005) \\
AntiBERTa2-CSSP & 0.998 (0.000) & 0.998 (0.000) & 0.996 (0.004) & \textbf{0.997 (0.002)} \\
IgBert & 0.993 (0.008) & 0.994 (0.006) & 0.986 (0.021) & 0.990 (0.013) \\
NanoBERT & 0.988 (0.001) & 0.990 (0.001) & 0.947 (0.005) & 0.967 (0.002) \\
VHHBERT & 0.986 (0.002) & 0.989 (0.001) & 0.939 (0.001) & 0.963 (0.001) \\
\bottomrule
\end{tabular}
\end{adjustbox}
\end{table}

\end{document}